\documentclass[sigconf]{acmart}
\AtBeginDocument{%
  \providecommand\BibTeX{{%
    \normalfont B\kern-0.5em{\scshape i\kern-0.25em b}\kern-0.8em\TeX}}}

\copyrightyear{2022}
\acmYear{2022}
\setcopyright{acmcopyright}\acmConference[CIKM '22]{Proceedings of the 31st ACM International Conference on Information and Knowledge Management}{October 17--21, 2022}{Atlanta, GA, USA}
\acmBooktitle{Proceedings of the 31st ACM International Conference on Information and Knowledge Management (CIKM '22), October 17--21, 2022, Atlanta, GA, USA}
\acmPrice{15.00}
\acmDOI{10.1145/3511808.3557427}
\acmISBN{978-1-4503-9236-5/22/10}
\usepackage{graphicx}
\usepackage{pifont}
\usepackage{amsmath}
\usepackage{hhline}
\usepackage{color}
\usepackage{diagbox}
\usepackage{graphicx}
\usepackage{array}
\usepackage{enumitem}
\usepackage{multirow}
\usepackage{ragged2e}
\usepackage{vcell}
\usepackage{float}
\usepackage{subcaption}
\usepackage{caption}
\usepackage{algorithm}
\usepackage{algpseudocode}
\usepackage{bm}

\definecolor{dg}{rgb}{0.82, 0.1, 0.26}

\DeclareMathOperator*{\argmax}{argmax}

\begin{document}

\title{Reinforced Continual Learning for Graphs}

\author{Appan Rakaraddi}
\affiliation{%
  \institution{Nanyang Technological University}
  \country{Singapore}
  }
\email{appan001@e.ntu.edu.sg}

\author{Lam Siew Kei}
\affiliation{%
  \institution{Nanyang Technological University}
  \country{Singapore}
}
\email{assklam@ntu.edu.sg}

\author{Mahardhika Pratama}
\affiliation{
 \institution{University of South Australia}
 \country{Australia}
 }
\email{dhika.pratama@unisa.edu.au}

\author{Marcus de Carvalho}
\affiliation{%
  \institution{Nanyang Technological University}
  \country{Singapore}
  }
\email{marcus.decarvalho@ntu.edu.sg}

% \author{Anonymous authors}

\begin{abstract}
Graph Neural Networks (GNNs) have become the backbone for a myriad of tasks pertaining to graphs and similar topological data structures. While many works have been established in domains related to node and graph classification/regression tasks, they mostly deal with a single task. Continual learning on graphs is largely unexplored and existing graph continual learning approaches are limited to the task-incremental learning scenarios. This paper proposes a graph continual learning strategy that combines the architecture-based and memory-based approaches. The structural learning strategy is driven by reinforcement learning, where a controller network is trained in such a way to determine an optimal number of nodes to be added/pruned from the base network when new tasks are observed, thus assuring sufficient network capacities. The parameter learning strategy is underpinned by the concept of Dark Experience replay method to cope with the catastrophic forgetting problem. Our approach is numerically validated with several graph continual learning benchmark problems in both task-incremental learning and class-incremental learning settings. Compared to recently published works, our approach demonstrates improved performance in both the settings. The implementation code can be found at \url{https://github.com/codexhammer/gcl}.
\end{abstract}




\maketitle

\section{Introduction}

The efficacy of Graph Neural Networks (GNNs) in deep learning has been proven for non-grid / non-Euclidean data where traditional deep learning architectures like CNNs fail to perform to the desirable levels. Different GNN models are specifically developed to tackle tasks related to node classification and regression, link prediction and graph classification-related tasks. Some of the applications with GNN models are citation network node classification \cite{gcn,sage}, graph matching \cite{matching}, graph clustering \cite{cluster}, regression tasks \cite{eigen} etc. The most widely-used form of GNNs are the Message-Passing GNNs (MP-GNNs), which aggregate and combine the information from the neighbouring nodes to form a hidden node feature. The most popular MP-GNNs for these aforementioned tasks are GCN \cite{gcn}, GAT \cite{gat}, GraphSAGE \cite{sage}, etc. 
\par
Despite the advances in different GNN architectures \cite{gcn,gat,sage}, these works focus on a single task problem. \textit{Continual learning} aims to learn a model handling a sequence of different tasks and the model should be able to answer any queries of already seen tasks without any bias towards a particular task. The major drawback in the continual learning scenario is the problem of catastrophic forgetting \cite{french1999catastrophic,ratcliff1990connectionist,mccloskey1989catastrophic}, where a model when trained on a new task tends to forget or perform poorly when tested on the previously trained tasks. This is because previously valid parameters are over-written when learning new tasks and the model forgets the previously learned tasks. Different methods are proposed to overcome the catastrophic forgetting problem and are mainly classified into 3 categories: memory-based methods \cite{rebuffi2017icarl,rolnick2019experience,isele2018selective,chaudhry2019continual},  regularization-based methods \cite{li2017learning,kirkpatrick2017overcoming,rannen2017encoder} and  architecture-based methods \cite{rcl,rusu2016progressive,fernando2017pathnet,aljundi2017expert}. All these existing methods mainly focus on grid-based data like images. Direct applications of these methods to non-grid based data like graphs is not possible/may perform poorly due to the different topological structure of a graph in comparison to planar images.

  Consider the Cora graph dataset which has 7 node classes. In traditional GNN learning models, the model learns all the 7 classes simultaneously. But by drawing comparison to how humans learn tasks (i.e., sequentially), the model should also learn the classification on Cora dataset by training on only a fraction of the classes at a time per task (e.g., say 2 classes per task). In practice, the number of tasks is unknown and a model should be prepared to handle a possible long sequence of tasks with modest computational and memory burdens. We argue that this problem cannot be handled with a static network structure, as it risks having inadequate network parameters to learn new tasks without succumbing to the catastrophic forgetting problem. The structural learning strategy must be adopted to determine new resources to be integrated and old resources to be removed. Such approach is exemplified in  \textit{Progressive neural networks} \cite{rusu2016progressive} but it suffers from considerable network complexity because new network components are blindly added when observing a new task. Although a loss-based network growing strategy in \textit{Dynamically expandable networks} \cite{dynamic} to incrementally add new nodes can be deployed, such approach does not assure an optimal resource allocation. In addition, a growing strategy which is done one-by-one is deemed too slow to adapt to new environments. The concept of reinforcement learning for continual learning is first adopted in RCL \cite{rcl} to determine an optimal number of filters to be added when coping with new tasks. This approach so far has been applied to image-like data and not yet extended to non-grid based data.  
\par

We propose a graph continual learning approach called \textit{Graph Continual Learning} (GCL). GCL is underpinned by the structural learning-based reinforcement learning approach to evolve the network structure to deal with new tasks. Reinforcement learning concept is implemented to train a controller network determining optimal actions when observing new tasks, i.e., new nodes to add or old nodes to prune. That is, our approach comprises a controller network called the \textit{Reinforcement Learning based Controller} (RLC) and the base trainable network called the \textit{Child Network} (CN).
\begin{enumerate}    

\item \textit{Reinforcement Learning based Controller} (RLC) network is a LSTM network trained via Reinforcement Learning. Its role is to determine the number of hidden features to be added or deleted in the GNN framework. 

\item \textit{Child Network} (CN) is constructed as a dynamically evolvable GNN framework where the hidden-layer node features are added and deleted as controlled by the RLC. This structural learning strategy assures a compact network structure to be developed when dealing with a sequence of different tasks.
\end{enumerate}

The parameter learning strategy of GCL is also driven by a knowledge distillation-based experience replay technique which replays the older samples stored in the memory from the previous tasks. That is, the stored knowledge of the memory samples are referenced to prevent the catastrophic forgetting problem in the subsequent tasks. To ensure this, we follow a formerly introduced method for images called dark experience replay (DER) \cite{der}, which deploys both model logits and classes for knowledge distillation and extend the concept to graph data. 

%A similar method with Reinforcement learning has previously been introduced for grid-like structures (e.g., images) \cite{rcl}, but we are the first to introduce such a method on non-grid structures coupled with Experience Replay for Continual Learning. 

There have been previous works focusing on continual learning on graphs with GNNs, which train the model in task-incremental settings with different GNN architectures~\cite{gnncl,cler}. \citeauthor{gnncl} \cite{gnncl} introduced a model called TWP that leverages the graph topological structure by regularization parameters. \citeauthor{cler} \cite{cler} introduced a model called ER-GNN that uses node experience replay for graph continual learning. 
These models are limited in their approach only to the task-incremental learning setting and have not been evaluated for the class-incremental learning. 
Our approach GCL, handles both task-incremental and class-incremental learning settings by introducing the concept of structural learning to assure the optimal network structure combined with the experience replay approach to avoid the catastrophic forgetting problem. \\

Our contributions are summarized as follows:
\begin{itemize}
    \item[\ding{51}] We develop a novel Graph Continual Learning (GCL) framework for continual learning on graphs. GCL combines the structure-based approach and the memory-based approach to cope with streaming tasks without catastrophic forgetting. 
    \item[\ding{51}] We propose the structural learning-based reinforcement learning approach with a controller network to predict optimal actions when observing new tasks. That is, it determines an optimal number of new nodes to be added or an optimal number of old nodes to be removed. 
    \item[\ding{51}] GCL is generalized framework applicable for both the task-incremental and class-incremental learning problems. It advances the existing approaches for continual learning on graphs, which are currently limited to the task-incremental learning problem. Note that the class-incremental learning problem is more challenging than the task-incremental learning problem because of the absence of task IDs in the former, making it harder to identify the task no. in which a particular class belongs to.
    \item[\ding{51}] We extensively conduct experimental study to demonstrate the effectiveness of our model over other state-of-the-art methods. We report the advantage of our approach compared to recently published works where ours achieves improved accuracy.  
\end{itemize}

\section{Related work}
\subsection{Graph Neural Networks}
GNNs \cite{scarselli,gcn,gat,sage} are becoming increasingly popular for deep learning on data pertaining to graph topology (or other similar non-Euclidean data) due to their better node representational learning ability. These abilities have been demonstrated across different tasks~\cite{cluster,eigen}, where the GNN takes in the higher-dimensional node features (and edge features, if present) and maps them to low-dimensional vector embedding. These embedding are used for solving different tasks. The GNNs are classified based on the method of aggregation which is mainly through spectral and spatial embedding \cite{zhou2020graph}. The spectral methods \cite{gcn,defferrard2016convolutional} learn the graph representation by Fourier transform of the graph spectral data; in contrast to spatial methods which follow the simple spatial aggregation of features from the neighbours \cite{sage,gat}. 
The new class of GNNs that are gaining prominence are WL-GNNs due to the limitation posed by expressive power of MP-GNNs in distinguishing graph isomorphism \cite{xu2018powerful,morris2019weisfeiler,Leman2018THERO}.
Most of these works are devised for a single learning task and are not directly applicable to the continual learning problem.

\subsection{Continual learning}
Continual learning is a longstanding problem in machine learning to handle a sequence of different tasks \cite{kirkpatrick2017overcoming,rebuffi2017icarl,french1999catastrophic,cl-review,andri,wei}. A continual learner accumulates knowledge from already seen tasks to minimize the catastrophic forgetting, thus gaining improved intelligence as it increasingly learns newer tasks. Existing approaches are grouped into three categories \cite{Parisi2019ContinualLL}: regularization-based approach, architecture-based approach and memory-based/ generative approach. The regularization-based methods \cite{li2017learning,kirkpatrick2017overcoming,rannen2017encoder} adopt an additional regularization term on top of the loss function to prevent important parameters from deviations. Although this approach is simple to implement, it does not scale well for a large-scale problem or dataset because of difficulties in finding the overlapping regions across all tasks. 
% In addition, this method is hardly applicable for the class-incremental learning problem with a single-head network configuration. 
The memory-based/ generative approach \cite{rebuffi2017icarl,rolnick2019experience,isele2018selective,chaudhry2019continual,shin2017continual} stores a small subset of old samples of previous tasks into a memory or rather generate samples of pseudo-data of the previously learned tasks to eliminate the  memory dependency. These memory-stored/generated samples are interleaved with the current-task samples for experience replay to minimize the catastrophic forgetting problem. This approach offers improved performances compared to other two approaches but with extra memory/generative costs. The architecture-based approach relies on the idea of network expansion when handling new tasks. New network components bring free network parameters which can be allocated to new tasks thus adapting quickly to new tasks. The catastrophic forgetting problem is addressed by isolating old network parameters. As with the regularization-based approach, this approach depends on the task-IDs, thus being mostly impractical for the class-incremental learning setting. Although most continual learning approaches are crafted for grid-based data such as images, some approaches for continual learning on graphs are proposed in \cite{gnncl,cler}, where \cite{gnncl} is based on the regularization-based approach while \cite{cler} is developed from the memory-based approach. Both approaches have not been evaluated for the class-incremental learning setting. GCL in this paper offers an alternative amalgamated approach combining the architecture-based approach and the memory-based approach, thereby handling the class-incremental learning situation as well as the task-incremental learning problem. The structural learning strategy assures sufficient and compact network structures while the parameter learning step with the help of a tiny memory avoids the catastrophic forgetting problem. No parameter isolation strategy is implemented in our approach because network parameters are indeed shareable across related tasks, i.e., association of specific parameters to specific tasks reduces flexibility.    
%The continual learning problem is formulated into three different learning scenarios \cite{scenario3}: Task-incremental (Task-IL), domain-incremental (Domain-IL) and class-incremental (Class-IL) settings. %These scenarios primary difference lies in whether during the learning, the task-ids are available or not and if it is not, then whether the model has to explicitly identify the identity of the task that is to be solved. The Task-IL is usually done with multi-headed layout( i.e., separate classifier/output layer for each task) whereas Class-IL is done with single-headed split.
%Our work focuses on both task-incremental and class-incremental settings.

% Expand on different methods

Our work improves upon the existing methods for continual learning for graphs, which are bolstered by the experimental results. To the best of our knowledge, we are the first to propose a consolidated model that can perform continual learning on graphs under both task-incremental and class-incremental settings. Also, our work proposes both addition and deletion of nodes to a graph neural network layer which increases the flexibility of the structural learning unlike the former works which only add nodes to the image-data learning neural network layer.

\subsection{Structural Learning-based Reinforcement Learning}
The use of reinforcement learning based controller network was introduced for Neural Architecture Search (NAS) \cite{nas,enas,pnas} to design a neural network from decisively selecting components from a predefined search space and determine the optimal parameters of components like CNNs, RNNs etc. This concept was later extended to graphs in GraphNAS \cite{graphnas} to decide the number of hidden layer features, variants of MP-GNNs, number of layers in the network etc. The first work to leverage RL-based controller for continual learning scenario was done for RCL \cite{rcl} but its scope was focused/limited to Euclidean-based data like images.  Our work extends this domain to graph structures to tackle the continual learning problem. The structural learning-based RL strategy is adopted in GCL because it allows an optimal action to be carried out and optimal extra parameters to be added compared to other approaches \cite{dynamic,Pratama2021UnsupervisedCL}. %The RL controller decides the CNN parameters like filter size, stride length etc. while expanding the base network, but as mentioned earlier, these methods focus on grid-based data while our work is the first to introduce such a controller for graphs for continual learning.

\begin{figure*}
\captionsetup[subfigure]{aboveskip=0pt,belowskip=15pt}
\begin{subfigure}{0.95\textwidth}
\vspace{2\baselineskip}
\includegraphics[scale=0.45]{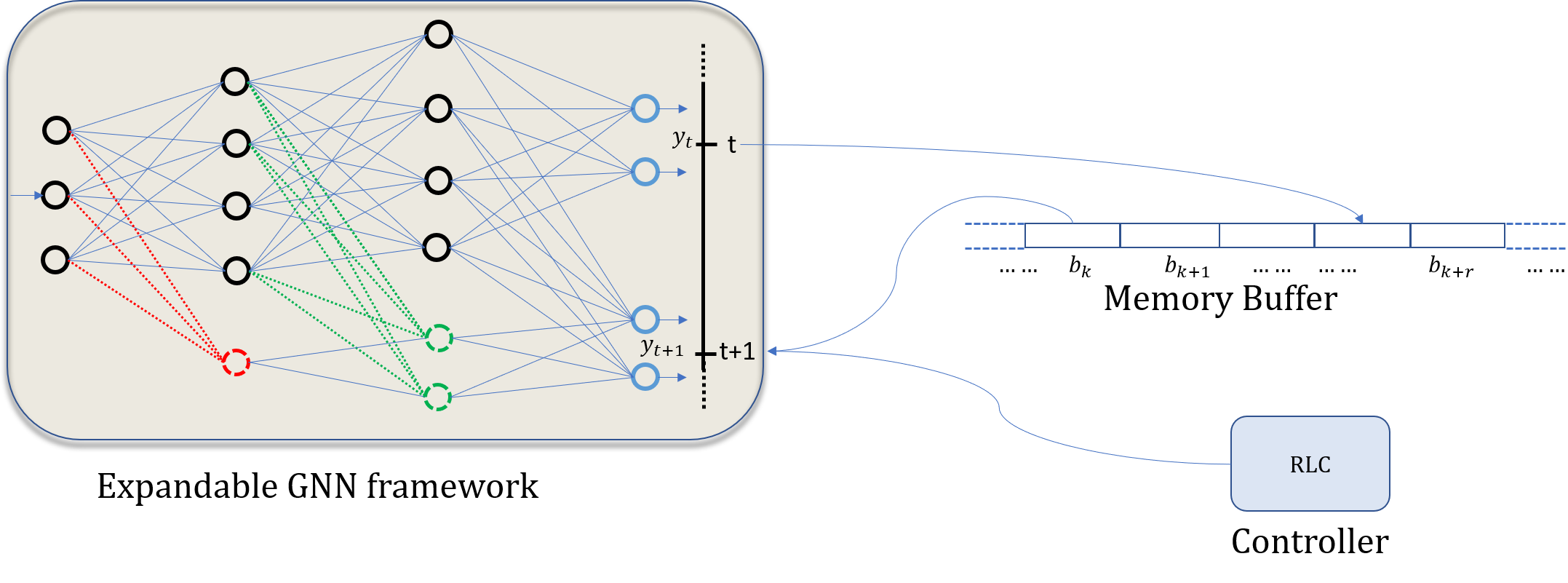}
\caption{GCL framework}
\label{fig:above}
\end{subfigure}

\begin{subfigure}{0.5\textwidth}
\includegraphics[scale=0.5]{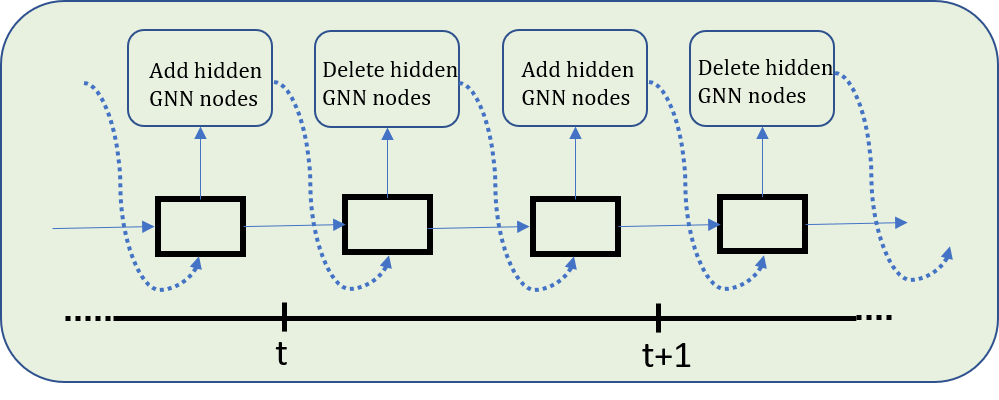}
\caption{RLC network}
\label{fig:below}
\end{subfigure}

    \caption{The outputs of each task, raw data and task-id $\textbf{t}$ are stored in the fixed-length buffer and retrieved in the subsequent tasks via reservoir sampling for Experience Replay. RLC network comprehends the current GNN state for structural learning to dynamically expand it by selecting the optimal actions from the search space for the next task-id $\textbf{t+1}$.}
    \label{fig:gcl}
\end{figure*}

\section{Problem formulation}

Let $\textbf{T} = \{t_i \ ; i \in [1,\mathcal{T}]\}$ be the set of $\mathcal{T}$ number of sequential tasks with the corresponding task-ids $t_i$. Each task $t_i \equiv (\mathcal{X}^{(t_i)}, \mathcal{Y}^{(t_i)})$ comprises of graph data and the nodes classes which is segregated into train, validation and test datasets. 
% Also, let $f_t$ represent the neural network state at task $t$ with parameters $\theta_t$ and $\mathcal{L}oss(.)$ be the loss function.

The aim of our method is to learn the node classification task from the data arriving sequentially with strict task boundaries \cite{scenario3} under 2 different settings: 
\begin{enumerate}
    \item Given the task-id, identify the class of a node (Task-Incremental setting).
    \item With task-id absent, identify the class of a node  (Class-Incremental setting).
\end{enumerate}

To accomplish the above, a Reinforcement Learning based structural learning GNN framework with Experience Replay called \textit{Graph Continual Learning} (GCL) is introduced. The \textit{Reinforcement Learning Controller} (RLC) dynamically evolves the \textit{Child Network} (CN) based on the network state from a pre-defined set of actions. To further prevent the catastrophic forgetting problem, a customised Experience Replay method is introduced on graph learning. We discuss these in detail as follows.

% Under defn. section  ---

% \subsubsection{Task-incremental setting:}
% The task-ids are available during the training and evaluation periods under this scenario. The most common way of learning is via multi-headed outputs where a head is active only during the training/testing of the task-classes in that head.
% Hence, the aim of task-incremental \cite{cl-survey} setting  i.e.,

% \begin{equation*}
% \hat{\theta} = \argmin_{\theta}    \sum_{t=1}^{\mathcal{T}} \mathbb{E}_{  (  \mathcal{X}^{(t)}, \mathcal{Y}^{(t)})   } \mathcal{L}oss  \left (f_t   (\mathcal{X}^{(t)}  ; \theta_t) ,  \mathcal{Y}^{(t)} \right)
% \end{equation*}

% \subsubsection{Class-incremental setting:} The sequential learning happens similar to as in task-incremental setting, but unlike the former, the task-ids are unavailable for the classification purposes \cite{masana2020class} in this case. Objectively, this makes class-incremental setting a more challenging task compared to the former. If $N_t$ represents all the tasks seen until task $t$, then the objective of class-incremental setting is defined by:

% \begin{equation*}
% \hat{\theta} = \argmin_{\theta}    \sum_{t=1}^{\mathcal{T}} \mathbb{E}_{  (  \mathcal{X}^{(N_t)}, \mathcal{Y}^{(N_t)})   } \mathcal{L}oss  \left (f_t   (\mathcal{X}^{(N_t)}  ; \theta_t) ,  \mathcal{Y}^{(N_t)} \right)
% \end{equation*}

\section{Preliminaries}

% Problem definition

We first briefly recap the message-passing nature of the GNNs. These GNN models are typically trained under single-task, all-node classification scenario which is inconsistent with the real-world scenarios. This evokes the the continual learning conundrum and we establish the problem statement under this presumption.

\subsection{Graph Neural Networks}

Given a graph $\mathcal{G}(V,E)$ and node feature vector $\mathcal{F}=\{f_i, \forall i \in V \}$, the node $u$'s representation in the $k$-th iteration is defined by:

\begin{equation}\label{eq:gnn}
  \textbf{h}_u^{(k)} =   \text{{{\footnotesize UPDATE}}}\left( \textbf{h}_u^{(k-1)}, 
    \text{\footnotesize {AGGREGATE}} \left( \{ \textbf{h}_v^{(k-1)}, \  \forall v \in   \mathcal{N}(u)\} \right)  \right) 
\end{equation}

where the {\footnotesize AGGREGATE} function gathers information from the neighbouring nodes of $u$ (i.e., $\mathcal{N}(u)$) and {\footnotesize UPDATE} function combines the node $u$'s former representation/feature vector to generate a newer one. The initial representation of a node at $k=0$ is its feature vector i.e., $\textbf{h}_u^{(0)}=f_u, \ \forall u\in V$. Equation \ref{eq:gnn} gives the generalized closed-form expression of multitude of different flavours of GNNs \cite{gat,gcn, sage}.

% \begin{definition}[Node classification]
% The classifier network $f(.)$ aims to accurately map the nodes to their respective classes upon the sequential arrival of the tasks in \textbf{T}.
% \end{definition}

\section{Methodology}

% In this section, we describe the architecture and the working of our model. 

% Change here

\subsection{Reinforcement Learning based Controller (RLC)}

% Elaborate RLC

% Structure Learning \ Parameter learning

The motive behind using the Reinforcement Learning (RL) \cite{rl} based controller is that an agent learns the optimal policy $\pi (a_t|s_t)$ to decide upon the appropriate set of actions $a_t$ when it has landed on the state $s_t$. This is done by maximising the expected sum of future reward values $R_t = \sum_{k=0}^{\infty}\gamma^{k}r_{t+k+1}$. The constant $\gamma \in (0,1]$ is the attenuation factor which accounts for the diminishing reward based on how far the reward is in the future. Hence, the action-value function or the $Q-$function is the expected returns/rewards when choosing an action from the current state.
This is expressed as:

\begin{equation}
    Q_{\pi}(s_t,a_t) =  \mathrm{E}_{\pi}\left [ R_t| S=s_t,A=a_t  \right ]
\end{equation}

Suppose the \textit{Child Network} (CN) has $m$ number of dynamically evolvable hidden layers, it becomes a key factor to identify the optimal course of actions $a_{1:m}$ to be taken on each of these layers. Simple brute-forced approach  of traversing across all set of values is highly inefficient as the search grows exponentially with the inclusion of every new layer, thus rendering it a NP-Hard problem. So, there is the need to optimally decide the necessary actions at each hidden layer which is fulfilled by the RLC network.

The RLC consists of LTSM network which learns the optimal state policy $\pi$ to select the appropriate course of actions $a_{1:m}$ from the user-defined search space $\mathcal{S}$ on each trainable layer of the CN. The $m$ number of action values are generated by taking into account the inter-dependency among the different layers of the CN (since backpropagation involves gradient multiplied across the layers of the network). The controller generates these action values $a_{1:m}$ in the form of an embedded fixed-length string token as the output of the LSTM network.
The search space $\mathcal{S}$ consists of 2 different types of actions for each CN layer: 

\begin{itemize}
    \item Number of hidden layer features addition (\textbf{ADD})
    \item Number of hidden layer features deletion (\textbf{DEL})
\end{itemize}

Each of these actions have a discrete set of values it can attain to be passed into the LSTM network. Hence for every evolvable layer, the number of features to be added is the difference of action values of \textbf{ADD} and \textbf{DEL}. The controller is trained over a number of epochs to generate a string of action values at every epoch where every string of action values output by the LSTM network is used to derive a different CN architecture. Among the multitude of CN architectures derived for a task, each of these architectures is associated with a reward value. The architecture corresponding to the highest reward value for a task is chosen as the optimal structure for that task. We will now discuss how to obtain the architecture-deciding reward value.

% \begin{table} [h]
% \centering
% \caption{Search space $\mathcal{S}$}
% \resizebox{\linewidth}{!}{%
% \begin{tabular}{|>{\hspace{0pt}}m{0.664\linewidth}||>{\hspace{0pt}}m{0.255\linewidth}|} 
% \hline
%  & \textbf{Values} \\ 
% \hline
% \textbf{Features addition action values / \textbf{ADD}} & 9, 11, 15, 21, 29 \\
% \textbf{Features deletion action values / \textbf{DEL}} & 1, \ 2, \ 5, \ 8, \ 9 \\
% \hline
% \end{tabular}
% }
% \end{table}

\subsubsection{Reward function.}  For training the CN, the data is divided between training (\textit{tr}), validation (\textit{vl}) and testing (\textit{ts}) sets. The reward function $R$, which is designed to tune the controller is based on the CN validation dataset average accuracy value. Hence, optimal action values $\hat{a}$ determined by the reward function $R$ in the state $s_t$ can be expressed as:

\begin{equation}
    \hat{a}_{1:m} = \argmax_{a \in \mathcal{S}} \mathbb{E}[ R(a_{1:m},s_t) ]
\end{equation}

The optimal actions $\hat{a}_{1:m}$ are used to dynamically evolve the CN which is trained as a normal deep learning framework, independent from the controller.
But since the reward function \textit{R} is non-differentiable, a policy-based approximation method called REINFORCE \cite{reinforce} algorithm is used for updating the RLC parameters $\theta_c$. This is defined by:

\begin{equation}
\label{eq:rlc}
    \nabla \mathbb{E}_{\pi}[R_{\theta_c}] = \mathbb{E} \left [
                \sum_{a_{1:m}} \nabla_{\theta_c} \ln \pi(a_t|s_t;\theta _c) \left( R(a_{1:m},s_t)-V(s_t)  \right) \right ]
\end{equation}

where $V(s_t)$ is the state-value function defined by $V(s_t) = \mathbb{E}[R_t|S=s_t]$ which is the moving target. In our work, we consider the moving target value as the average of the past architecture rewards by simple Monte-Carlo approximation.

% So at each step, the controller needs to maximise the expected value of the reward function (i.e., $\mathbb{E}[R]$) value of the CN  by selecting the appropriate search action values $\hat{a}$ i.e.,

\subsection{Child Network (CN)}
The CN is a multi-layered  cascaded GNN framework with $m$ dynamically evolvable hidden layers controlled by the RLC. Note that since the controller actions only consist of addition/deletion of hidden layer features/nodes i.e., \textbf{ADD} or \textbf{DEL}, it relaxes the constraints of choosing a particular flavour of MP-GNN, thus making our framework universal across different MP-GNN variants. The loss function for the current task is defined by:

\begin{equation}
        \mathcal{L}oss_{t_c} = \min_{\theta_{t_c}} \mathcal{L}oss  \left (f_{t} 
        (\mathcal{X}^{(t_c)}  ; \theta_{t}) , \  \mathcal{Y}^{(t_c)} \right)
\end{equation}

where ${\theta_t}$ are the trainable Child Network/CN parameters at time instance $t$; and $f_t$ is the state of the CN at time instance $t$. In our work, we define $\mathcal{L}oss(.)$ as the Cross-Entropy Loss function.

Since the CN is evolving, only the previous network state weight matrices are incorporated and there is a loss in the gradient values every time the network changes shape (as the weights are not freezed to preserve the gradients). This phenomenon can cause catastrophic forgetting problem which needs to be dealt with by further additions to the GCL.
% \subsubsection{Task-incremental setting:}
% The classifier/logits layer of the neural network (GNN) has the output number of nodes equal to that of the number of classes in the dataset. But during training/testing periods, since we  are allowed the task-ids, only those weights of those nodes are optimized in the logits layer whose classes are present in the current task while freezing the rest of the node weights. This is coherence with the multi-head setting.

% % Just add to problem definition
% \begin{equation}
%     \mathcal{L}oss_{t_c} = \min_{\theta_t} \mathcal{L}oss  \left (f_t   (\mathcal{X}^{(t)}  ; \theta_t) ,  \mathcal{Y}^{(t)} \right)
% \end{equation}

% \subsubsection{Class-incremental setting:}

% All the outputs of the classifier/logits layer are optimized which are associated with the classes seen/present till the current task (inclusive). To make availability of the previous task-data while training on the current task, task-data is sampled and stored in the buffer memory for replay purposes (discussed later on). The replay    \newline
% We use Adam optimizer for weight optimization and reinitialize the optimizer every time network size changes, thus dropping the previous gradients. Note that we adopt this strategy so that the entire model is retrained at every task for better optimization and complete utilization of the entire network, instead of optimizing only the expanded part of the network as has been done in the previous works \cite{rcl,den}.
\subsubsection{Experience Replay:}
% Be more elaborative about the terms

To further suppress the catastrophic forgetting, we introduce an additional \textit{Experience Replay} (ER) module that stores and replays data from the preceding tasks. Our work takes inspiration from the
Experience Replay previously introduced to deal with images referred to as \textit{Dark Experience Replay}++ \cite{der} and extends the same to the graph data.

For replaying the previous tasks, we leverage a fixed size buffer where each block of the buffer memory consists of the CN outputs/logits ($l_t$) and the subgraph data sampled along every epoch (along with its node labels i.e., $(\mathcal{X}^{(t)} , \mathcal{Y}^{(t)}$) during the training process. The buffered data is replayed along with the current task-data in a joint loss-function to optimize the weights associated with the current task is $t_c$ and buffered tasks. The buffered tasks are sampled via \textit{reservoir sampling} and the joint-loss function is expressed as:

\begin{equation}
     \mathcal{L}oss_{net}  =  \mathcal{L}oss_{t_c} +   \alpha  \left |\left|  f_{t} ( \mathcal{X}^{(t_i)} ) -  l_{t_i}   \right |\right|_2 +   \beta  \mathcal{L}oss \left (  f_{t} ( \mathcal{X}^{(t_j)} )  - \mathcal{Y}^{(t_j)} \right ) 
\end{equation}

where $\alpha$ and $\beta$ are hyper-parameters, $t_i$ and $t_j$ are the task-ids of buffered tasks and $||.||_2$ denotes the MSE Loss function.

The complete algorithm is illustrated in Algorithm \ref{alg:gcl}.

\begin{algorithm}
\caption{GCL}\label{alg:gcl}
\begin{algorithmic}[1]
\Require $\{t_i \equiv (\mathcal{X}^{(t_i)}, \mathcal{Y}^{(t_i)}), \ \  \forall \  i \in [1,\mathcal{T}] \} $; number of controller steps \textit{max\_controller\_step}

\State Initialize \textit{CN} hidden layer sizes
\State Initialize \textit{RLC} 

\For{$t = 1,...,\mathcal{T}$}
    \If{$t=1$}
        \State Train and optimize \textit{CN}
    \Else 
    
\For{\textit{c} = 1,...,\textit{max\_controller\_step}}

\State Get actions $a_{1:m}$ from \textit{RLC}
\State Evolve and train \textit{CN}'s hidden layers based on $a_{1:m}$
\State Train \textit{RLC} based on the reward fn. in Equation (\ref{eq:rlc})

\EndFor
\State Select the best architecture based on the reward value
\State Re-train CN based on this architecture

\EndIf
\EndFor
\end{algorithmic}
\end{algorithm}

\section{Experiments}

We have performed experiments on a variety of datasets to demonstrate the effectiveness of our method for both task and class incremental settings. We have also conducted ablation study to breakdown the component-wise model accuracy.

\subsection{Datasets}
We have conducted experiments on four benchmark datasets:
\begin{enumerate}[label=\roman*)]
    \item \textit{Cora} dataset \cite{cora-citeseer}: For both task-incremental and  class-incremental  setting, the dataset is divided into 3 tasks with 2 classes per task.
    \item \textit{Citeseer} dataset \cite{cora-citeseer}: For both task-incremental and class-incremental setting, the dataset is divided into 3 tasks with 2 classes per task.
    \item \textit{Amazon Computers} dataset \cite{amazon}: For both task-incremental and class-incremental setting, the dataset is divided into 5 tasks with 2 classes per task.
    \item \textit{Corafull} dataset \cite{corafull}: For both task-incremental and class-incremental setting, the dataset is divided into 9 tasks with 5 classes per task. Note that even though this dataset has 70 classes, we have discarded the classes whose cardinality is less than 150.
\end{enumerate}
The dataset details and task-splitting are summarized in Table \ref{table:dataset}.

\subsection{Baselines}
To demonstrate the method effectiveness, we have compared with currently existing state-of-the-art methods. The baselines for Task-incremental settings include all the baselines used in Class-incremental settings along with some additional baselines. Below we list out the \textit{Common Baselines} (both Task-incremental and Class-incremental) and \textit{Task-exclusive Baselines} (Task-incremental only).

\subsection*{Common Baselines}

\subsubsection{GCN} \cite{gcn}.  A standard GCN uses the mean aggregator in Equation \ref{eq:gnn} for node hidden layer representation.

\subsubsection{GAT} \cite{gat}. GAT is an-isotropic MP-GNN by assigning attention coefficients to the neighbouring node feature vectors during the feature aggregation to generate the node embedding.

\subsubsection{GraphSAGE} \cite{sage}. GraphSAGE is inductive spatial aggregator that uses sample and aggregate method from the local neighbourhood of a node.

\begin{table}
\centering
\caption{Dataset properties and task-splitting}
\label{table:dataset}
\resizebox{\linewidth}{!}{%
\begin{tabular}{|l||c|c|c|c|} 
\hhline{|=:t:====|}
\diagbox{Properties}{Dataset} & \multicolumn{1}{l|}{Cora} & \multicolumn{1}{l|}{Citeseer} & \multicolumn{1}{l|}{Corafull} & Amazon Computers \\ 
\hhline{|=::====|}
No. of nodes & \textcolor[rgb]{0.251,0.251,0.251}{2,708} & 3,327 & \textcolor[rgb]{0.251,0.251,0.251}{19,793} & 13,752 \\
No. of edges & \textcolor[rgb]{0.251,0.251,0.251}{10,556} & \textcolor[rgb]{0.251,0.251,0.251}{9,104} & 126,842 & 491,722 \\
No. of features & \textcolor[rgb]{0.251,0.251,0.251}{1,433} & \textcolor[rgb]{0.251,0.251,0.251}{3,703} & \textcolor[rgb]{0.251,0.251,0.251}{8,710} & \textcolor[rgb]{0.251,0.251,0.251}{767} \\
No. of classes & \textcolor[rgb]{0.251,0.251,0.251}{7} & \textcolor[rgb]{0.251,0.251,0.251}{6} & \textcolor[rgb]{0.251,0.251,0.251}{70} & \textcolor[rgb]{0.251,0.251,0.251}{10} \\ 
\hhline{|=::====|}
No. of tasks & 3 & 3 & 9 & 5 \\
No. of classes per task & 2 & 2 & 5 & 2 \\
\hhline{|=:b:====|}
\end{tabular}
}
\end{table}

\subsection*{Task-exclusive Baselines}

% \subsubsection{ER-GNN} \cite{cler}. This method was proposed Experience Replay method on graphs for task-incremental setting under different Replay buffer strategy. More specifically, we use GAT as GNN combined with Experience Replay.  The comparison of this baseline is only done with the dataset used in their paper for fairer comparison.

\subsubsection{LWF} \cite{li2017learning}. This method performs knowledge distillation of the older tasks and uses joint loss function along with the regularization term to learn the newer tasks without forgetting the older ones.

\subsubsection{TWP} \cite{gnncl}. This method captures the graph topology and finds the crucial parameters for task-incremental learning.

\subsubsection{EWC} \cite{kirkpatrick2017overcoming}. The EWC method uses the Fisher information matrix to determine the parameter importance and penalizes the changes in them accordingly to alleviate the catastrophic forgetting problem.

\subsubsection{GEM} \cite{lopez2017gradient}. This method uses episodic memory $\mathcal{M}_t$ to store the subset of samples from current task example which is replayed later to avoid the forgetting problem.

We have not included ER-GNN \cite{cler} as a baseline due to the unavailability of the official code which restricts a fairer comparison with the other methods.

\begin{table*}
\centering
\caption{Comparison of different models on various datasets under Task-incremental setting. The best values are highlighted in red. Higher the AA value, better is the performance. Lower the AF value, better is the performance.}
\label{table:task}
\resizebox{\linewidth}{!}{%
\begin{tabular}{|c|cc|cc|cc|cc|} 
\hline
\multirow{2}{*}{\textbf{Task-incremental setting}} & \multicolumn{2}{c|}{Cora} & \multicolumn{2}{c|}{Citeseer} & \multicolumn{2}{c|}{Corafull} & \multicolumn{2}{c|}{Amazon Computers} \\ 
\cline{2-9}
 & AA$(\uparrow)$ & AF$(\downarrow)$ & AA$(\uparrow)$ & AF$(\downarrow)$ & AA$(\uparrow)$ & AF$(\downarrow)$ & AA$(\uparrow)$ & AF$(\downarrow)$ \\ 
\hhline{|=========|}
GCN & $73.6 \pm 14.3\%$ & $11.1 \pm 16.8\%$ & $61.0 \pm 3.5\%$ & $0.6 \pm 12.9 \%$ & $78.3 \pm 4.0\%$ & $11.1 \pm 3.5\%$ & $65.1 \pm 10.3\%$ & $20.6 \pm 5.2\%$ \\
GAT & $70.4\pm 14.1\%$ & $3.0\pm 5.7\%$ & $64.2\pm 6.6\%$ & $3.5\pm 7.1\%$ & $60.9\pm 9.4\%$ & $19.6\pm 8.7\%$ & $62.1\pm 13.8\%$ & $18.6\pm 7.1\%$ \\
GraphSAGE & $59.2\pm 8.7\%$ & $2.7\pm 6.9\%$ & $52.3 \pm 5.5\%$ & $\color{dg}\bm{0.5 \pm 1.9\%}$ & $76.2\pm 3.3\%$ & $9.8\pm 3.2\%$ & $80.1\pm 8.2\%$ & $22.0\pm 9.8\%$ \\
LWF & $82.3\pm 2.9\%$ & $16.8\pm 4.2\%$ & $73.7\pm 1.5\%$ & $10.1\pm 2.1\%$ & $51.4\pm 4.8\%$ & $45.7\pm 5.0\%$ & $90.0\pm 6.9\%$ & $9.0\pm 8.8\%$ \\
TWP & $92.2\pm 0.7\%$ & $1.8\pm 0.6\%$ & $78.7\pm 1.1\%$ & $1.9\pm 2.3\%$ & $87.5\pm 1.1\%$ & $4.4\pm 0.8\%$ & $91.1\pm 3.6\%$ & $1.6\pm 4.1\%$ \\
EWC & $91.6\pm 1.0\%$ & $3.2\pm 1.2\%$ & $78.2\pm 1.5\%$ & $2.9\pm 2.9\%$ & $85.1\pm 1.5\%$ & $7.6\pm 1.0\%$ & $94.4\pm 3.0\%$ & $3.1\pm 3.0\%$ \\
GEM & $82.0 \pm 1.9\%$ & $\color{dg}\bm{0.0 \pm 4.4\%}$ & $71.9\pm 3.6\%$ & $5.9\pm 4.3\%$ & $76.0\pm 2.3\%$ & $14.7\pm 2.5\%$ & $84.9\pm 5.4\%$ & $23.8\pm 5.0\%$ \\ 
\hhline{|=========|}
GCL-GCN & $\color{dg}\bm{94.3\pm 0.9\%}$ & $2.0\pm 1.4\%$ & $78.3\pm 2.5\%$ & $4.0\pm 4.4\%$ & $\color{dg}\bm{92.8\pm 0.3\%}$ & $\color{dg}\bm{0.6\pm 1.0\%}$ & $94.9\pm 1.7\%$ & $\color{dg}\bm{0.1\pm 2.8\%}$ \\
GCL-GAT & $92.5\pm 1.4\%$ & $4.8\pm 1.7\%$ & $\color{dg}\bm{80.2\pm 1.9\%}$ & $1.7\pm 2.3\%$ & $89.8\pm 2.5\%$ & $2.3\pm 4.0\%$ & $97.2\pm 1.1\%$ & $3.3\pm 1.5\%$ \\
GCL-GraphSAGE & $91.3\pm 1.8\%$ & $15.2\pm 10.3\%$ & $79.3\pm 1.7\%$ & $3.8\pm 10.0\%$ & $87.5\pm 0.3\%$ & $6.2\pm 0.5\%$ & $\color{dg}\bm{97.6\pm 0.6\%}$ & $0.3\pm 0.5\%$ \\
\hline
\end{tabular}
}
\end{table*}

\subsection{Comparison Metrics}
To demonstrate the effectiveness of our method, two measures have been adopted: Average accuracy (AA) and Average Forgetting (AF) \cite{lopez2017gradient} in both task-incremental and class-incremental settings. A lower triangular matrix called the \textbf{R}-matrix, $R \in \mathbb{R}^{\mathcal{T}\times \mathcal{T}}$ ($\mathcal{T}$ is the number of tasks) is constructed where $R_{ij}$ is the accuracy value on the test dataset of task-$j$ after training on task-$i$.
The AA value is the mean accuracy across all the learned tasks on the test dataset or the mean of the last row of the $R$ matrix. Higher AA value signifies a better the model performance. The AF value determines the performance drop across the previous tasks when training on the current task and lower AF value signifies better model performance.

\subsection{Implementation details}
The LSTM-based RLC controller is set to train for maximum of 4 steps with batch size of 64 and each step constitutes 250 training epochs of the CN model on 8GB Nvidia GPU. The $\alpha$ and $\beta$ parameters in the Experience Replay are both set to 0.5 with a buffer memory slots of 1000 units. The CN has 2 hidden dynamically evolvable GNN layers with the initial number of hidden-layer nodes set to 20 on both the layers. Adam's optimizer is used for optimization of both the RLC and CN with learning rates set to $3.5\times10^{-4}$ and $5\times10^{-3}$ respectively. The code is implemented in PyTorch and the GNN framework is implemented with Pytorch-Geometric library \cite{pyg}.

The GCL framework is implemented in 3 evolvable MP-GNN variants. Each of these variants is tested for both task-incremental and class-incremental settings. For GCL-GAT, 2 attention heads are being used and the outputs are averaged instead of the more conventional concatenation operation. To accommodate the experience replay, the graphs are sub-sampled to generate smaller subgraphs which are pushed into the buffer at every epoch. We tested our model for 5 trials with consecutive seeds with the initial seed value set to 123.

\begin{figure}
    \centering
    \includegraphics[width=\columnwidth]{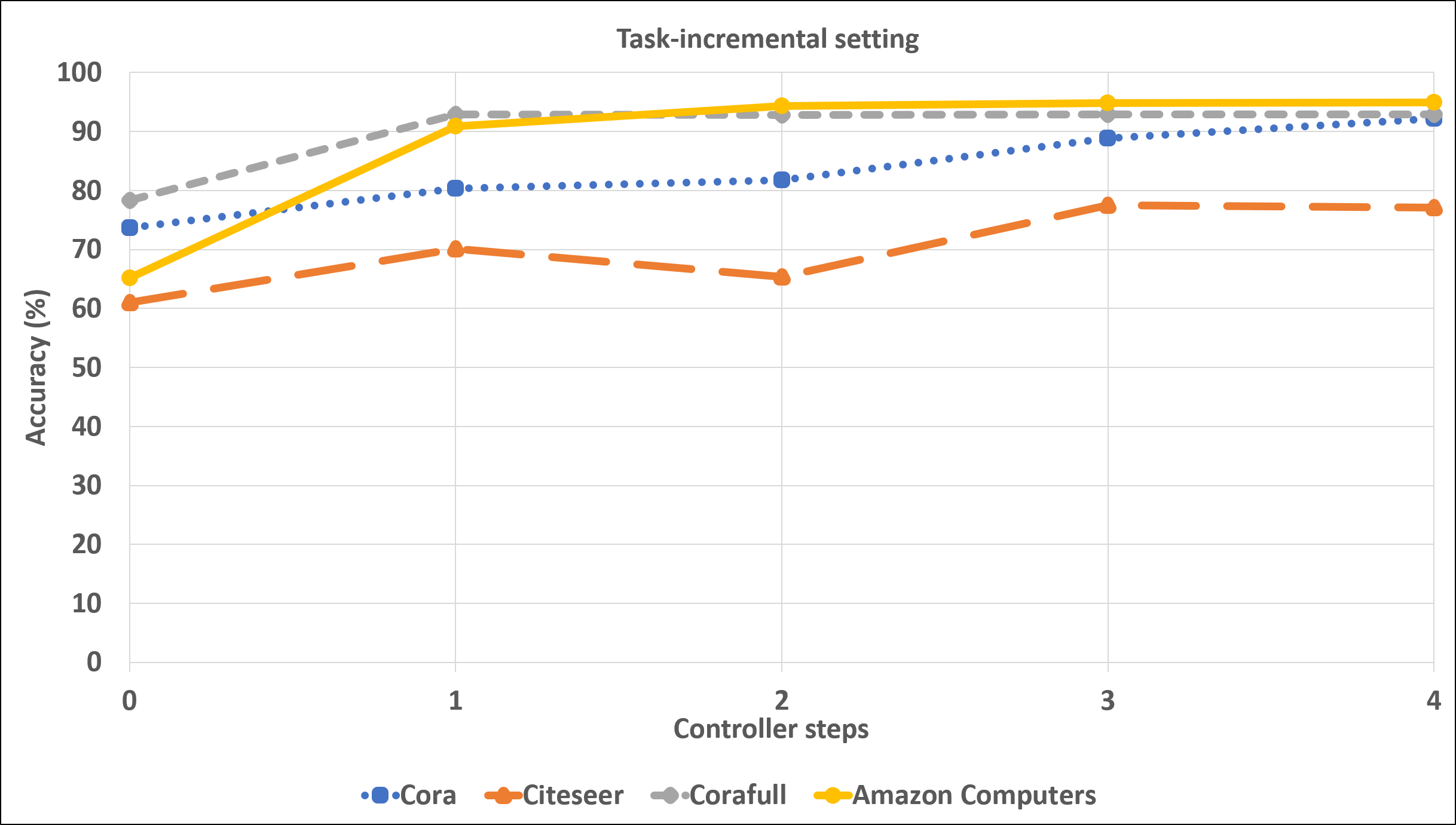}
    \caption{Variation in Average Accuracy in GCL-GCN with different number of RLC steps in Task-incremental setting on different datasets.}
    \label{fig:controller_t}
\end{figure}

\begin{table*}
\centering
\caption{Comparison of different models on various datasets under Class-incremental setting. The best values are highlighted in red. Higher the AA value, better is the performance. Lower the AF value, better is the performance.}
\label{table:class}
\resizebox{\linewidth}{!}{%
\begin{tabular}{|c|cc|cc|cc|cc|} 
\hline
\multirow{2}{*}{\textbf{Class-incremental setting}} & \multicolumn{2}{c|}{Cora} & \multicolumn{2}{c|}{Citeseer} & \multicolumn{2}{c|}{Corafull} & \multicolumn{2}{c|}{Amazon Computers} \\ 
\cline{2-9}
 & AA$(\uparrow)$ & AF$(\downarrow)$ & AA$(\uparrow)$ & AF$(\downarrow)$ & AA$(\uparrow)$ & AF$(\downarrow)$ & AA$(\uparrow)$ & AF$(\downarrow)$ \\ 
\hhline{|=========|}
GCN & $20.9\pm 7.8\%$ & $32.7\pm 23.0\%$ & $18.8\pm 3.9\%$ & $32.8\pm 16.2\%$ & $6.0\pm 1.5\%$ & $11.9\pm 3.0\%$ & $7.1\pm 8.0\%$ & $39.6\pm 18.9\%$ \\
GAT & $19.5\pm 8.7\%$ & $38.7\pm 20.5\%$ & $25.9\pm 9.8\%$ & $21.4\pm 17.0\%$ & $3.6\pm 0.8\%$ & $18.6\pm 5.2\%$ & $2.6\pm 2.9\%$ & $37.5\pm 18.1\%$ \\
GraphSAGE & $21.4\pm 7.4\%$ & $34.9\pm 13.0\%$ & $16.6\pm 0.5\%$ & $46.0\pm 19.5\%$ & $5.3\pm 2.6\%$ & $17.9\pm 5.6\%$ & $19.3\pm 0.3\%$ & $69.8\pm 24.3\%$ \\ 
\hhline{|=========|}
GCL-GCN & $\color{dg}\bm{55.7\pm 1.3\%}$ & $\color{dg}\bm{2.5\pm 14.3\%}$ & $\color{dg}\bm{36.5\pm 5.4\%}$ & $\color{dg}\bm{2.8\pm 8.2\%}$ & $25.3\pm 5.3\%$ & $2.1\pm 5.7\%$ & $37.2\pm 5.7\%$ & $3.7\pm 4.3\%$ \\
GCL-GAT & $54.2\pm 2.9\%$ & $7.2\pm 7.5\%$ & $34.5\pm 7.2\%$ & $11.7\pm 10.8\%$ & $20.3\pm 4.5\%$ & $2.1\pm 3.7\%$ & $22.8\pm 5.8\%$ & $\color{dg}\bm{3.1\pm 7.2\%}$ \\
GCL-GraphSAGE & $37.4\pm 3.5\%$ & $16.3\pm 6.6\%$ & $31.4\pm 9.4\%$ & $22.0\pm 13.5\%$ & $\color{dg} \bm{25.7\pm 2.9\%}$ & $\color{dg}\bm{0.7\pm 7.8\%}$ & $\color{dg}\bm{45.4\pm 10.2\%}$ & $6.3\pm 6.6\%$ \\
\hline
\end{tabular}
}
\end{table*}

\subsection{Task-incremental setting}
Table \ref{table:task} shows the AA and AF performance comparison for the 4 datasets with different methods. It can be clearly observed the performance is the worst for GCN, GAT and GraphSAGE since these methods do not use any techniques to alleviate the catastrophic forgetting problem. The methods TWP and EWC are closer in performance since both rely on the Fisher information matrix for parameter preserving.

\begin{figure}
    \centering
    \includegraphics[width=\columnwidth]{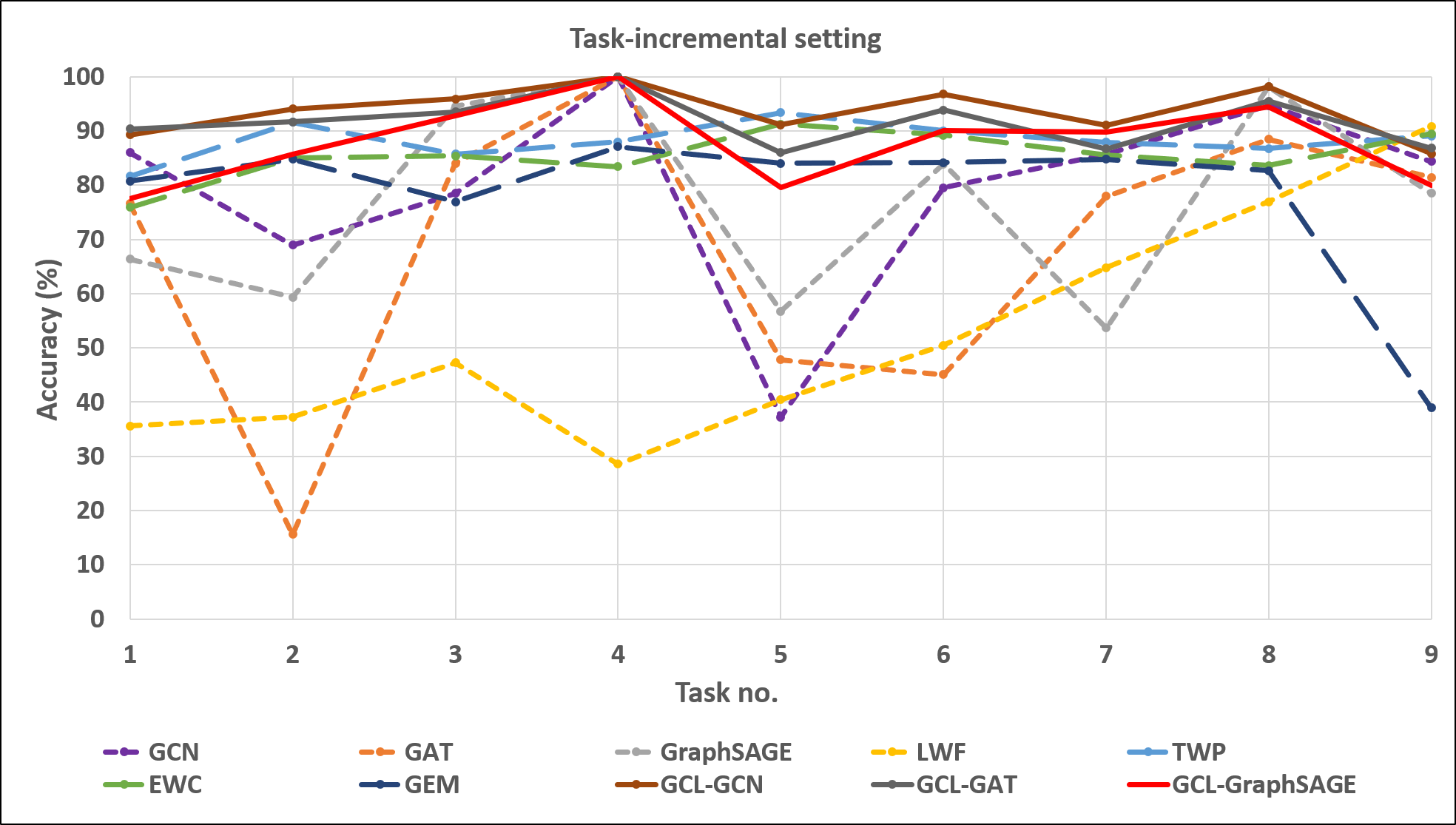}
    \caption{Accuracy on each task of Corafull dataset after training on 9 tasks with Task-incremental setting. The accuracy values of our models are in solid lines and rest of the baselines are in dashed lines. Our models have low fluctuations in accuracy values from task-to-task which indicates lowered forgetting problem.}
    \label{fig:tasks_corafull}
\end{figure}

\begin{figure}
    \centering
    \includegraphics[width=\columnwidth]{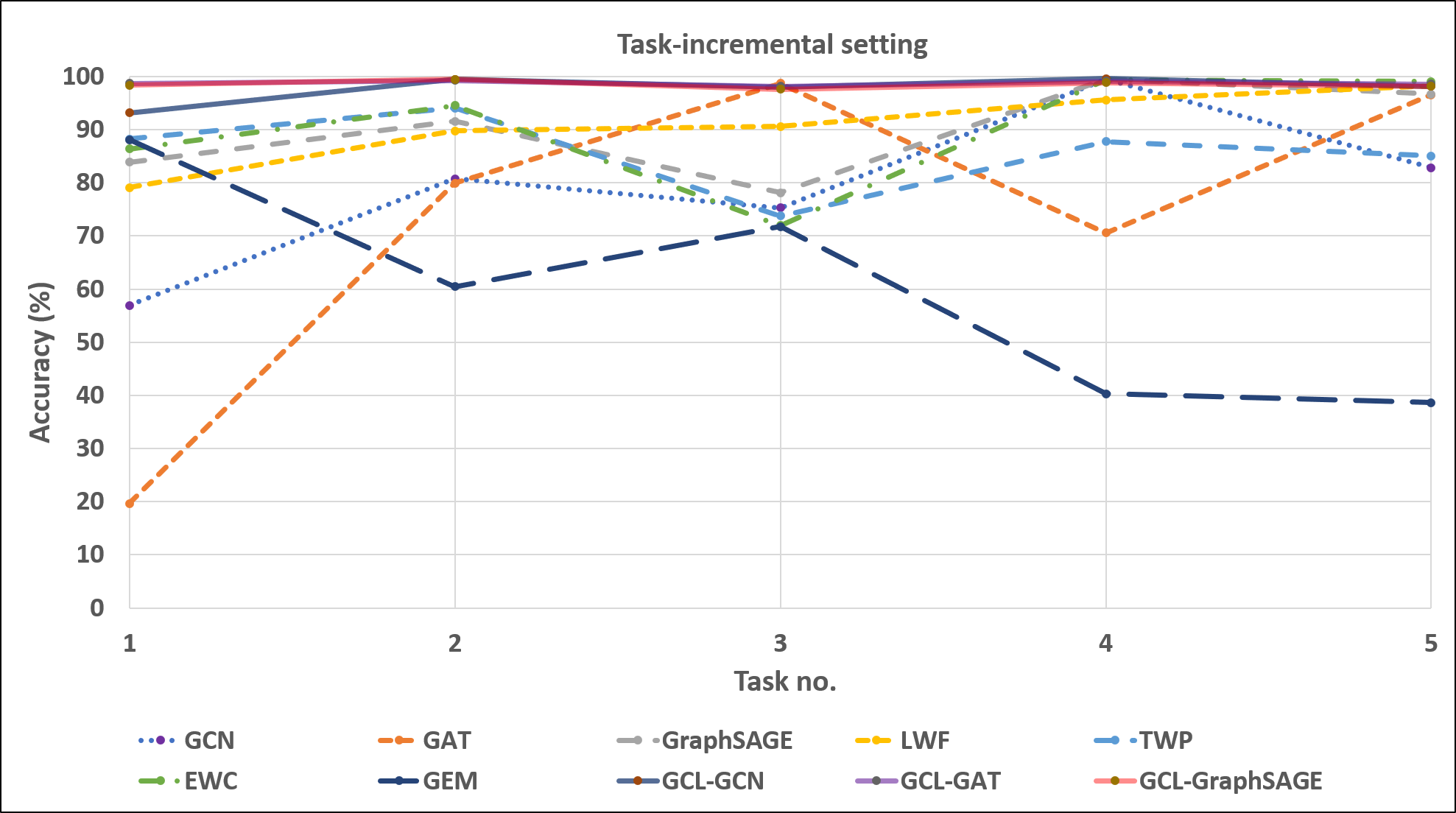}
    \caption{Accuracy on each task of Amazon Computers dataset after training on 5 tasks with Task-incremental setting. The accuracy values of our models are in solid lines and rest of the baselines are in dashed lines. Our models have low fluctuations in accuracy values from task-to-task which indicates lowered forgetting problem.}
    \label{fig:tasks_comp}
\end{figure}

As it can be observed in Figure \ref{fig:tasks_corafull} and Figure \ref{fig:tasks_comp}, our method GCL shows low perturbations in the accuracy values from task to task which is an indication of the low forgetting problem and better stability. Although TWP also depicts a similar phenomenon, it has relatively lower accuracy values across the tasks. It can be observed in Table \ref{table:task} that as the size of the dataset increases, our model performs much better relative to the others in both AA and AF. For example, Cora and Citeseer datasets are relatively smaller datasets with 3 tasks considered in our experiments with 2 classes per task. While our model mostly outperforms other baseline models on these datasets as well, it shows a higher performance margin as the size of the dataset increases (like in Corafull and Amazon Computers) and the number of tasks also increase. This coincides well with the real-life scenario where a neural network needs to learn and adapt to a large number of tasks with low amount of forgetting. Hence, this vouches for the GCL model scalability to large datasets with extensive large number of tasks and yet maintain low amount of forgetting with high average accuracy with low accuracy perturbations across individual tasks.

\subsection{Class-incremental setting}

\begin{figure}
    \centering
    \includegraphics[width=\columnwidth]{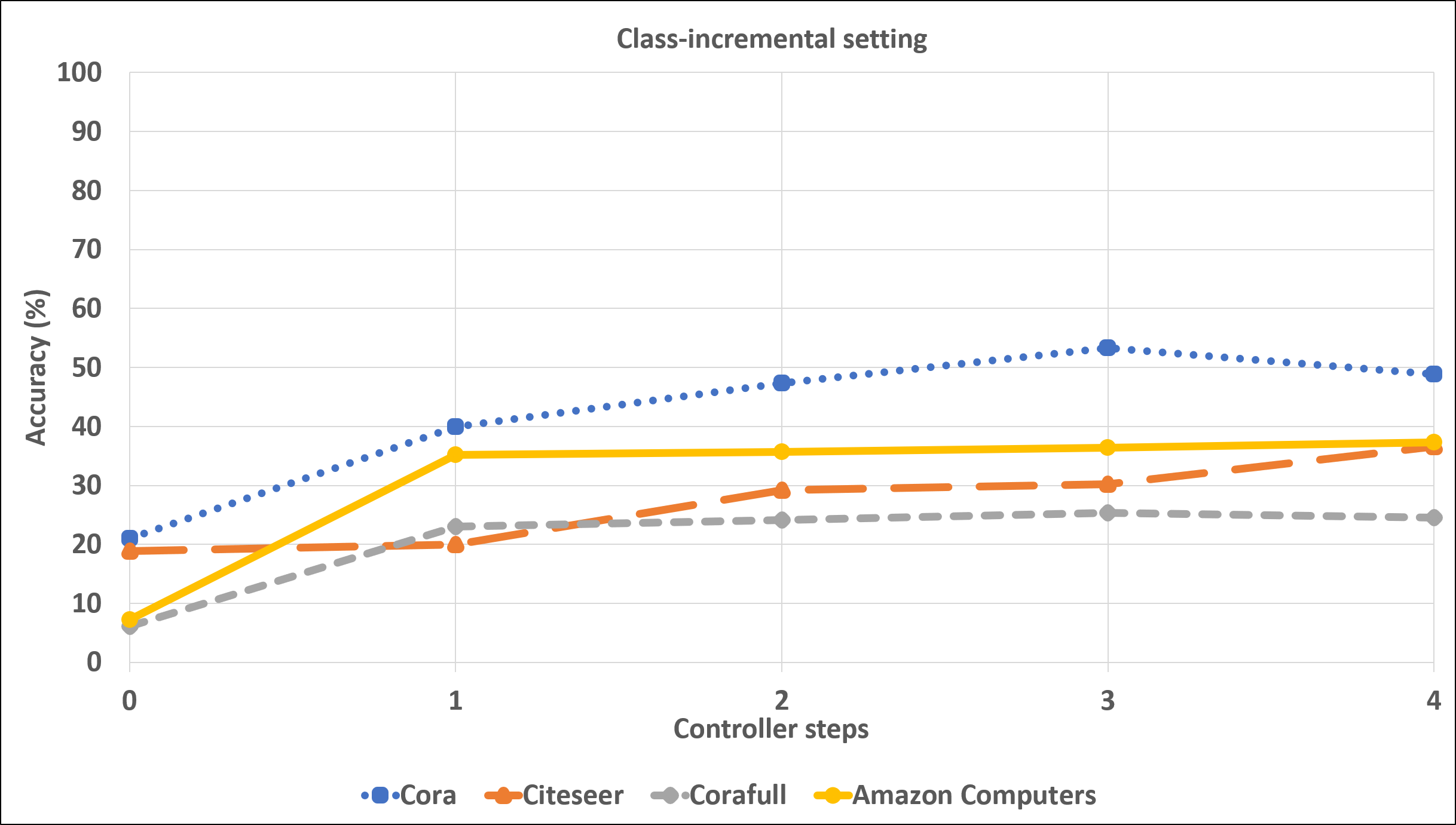}
    \caption{Variation in Average Accuracy in GCL-GCN with different number of RLC steps in Class-incremental setting on different datasets.}
    \label{fig:controller_c}
\end{figure}

Table \ref{table:class} shows the performance comparison of the common MP-GNN variants against their counterparts which are modified to alleviate the catastrophic forgetting problem in class-incremental setting. Our proposed methods consistently outperform the baselines in both the metrics across the datasets. It was observed that GCL-GAT did not have such a consistent superior performance. For example, in both Cora and Citeseer, GCL-GAT performed poorly in-terms of AA metric compared to the baselines, but performed much better in terms of AF metric. Figure \ref{fig:controller_c} depicts the variation in AA for GCL-GCN while trained with gradually increasing number of RLC steps from 0 (note that for 0 controller steps, we used simple GCN with no RLC or experience replay as our model) to 4 steps. It is interesting to note that we did not observe a strict increasing function between number of controller steps and AA. 

Although GCL-GCN performs the best in terms of AA on both the Cora and Citeseer datasets which have 3 tasks each, GCL-GraphSAGE takes the lead with Amazon Computers which has 5 tasks. GCL-GAT fared consistently with low forgetting as evidenced by the low AF values, albeit it under-performed a large margin in terms of AA in comparison to the other methods.

\subsection{Ablation Study}

\begin{table}[H]
\centering
\caption{Ablation study with Task-incremental setting on the components of GCL-GCN.}
\resizebox{\linewidth}{!}{%
\begin{tabular}{|c|cc|cc|} 
\hline
\multirow{2}{*}{\textbf{Task-incremental setting}} & \multicolumn{2}{c|}{Cora} & \multicolumn{2}{c|}{Citeseer} \\ 
\cline{2-5}
 & AA$(\uparrow)$ & AF$(\downarrow)$ & AA$(\uparrow)$ & AF$(\downarrow)$ \\ 
\hhline{|=====|}
ER + CN & $80.3\pm 14.9\%$ & $1.6\pm 1.5\%$ & $70.1\pm 8.4\%$ & $5.8\pm 7.7\%$ \\
RLC + CN & $81.7\pm 10.0\%$ & $9.6\pm 12.7\%$ & $65.2\pm 8.3\%$ & $8.5\pm 6.6\%$ \\ 
\hhline{|=====|}
GCL-GCN/ RLC + ER + CN & $90.2\pm 5.2\%$ & $1.5\pm 7.7\%$ & $77.9\pm 2.8\%$ & $5.5\pm 4.9\%$ \\
\hline
\end{tabular}
}
\label{tab:abl-task}
\end{table}

To demonstrate the effectiveness of the RLC and Experience Replay (ER) components, an ablation study was conducted with the individual components on \textit{Cora} and \textit{Citeseer} datasets using GCL-GCN in both task-incremental and class-incremental settings. In both the settings, we first verified the effectiveness of replaying the previously seen tasks with the ER component using the (ER + CN) connection-based model. Next, we verified the effectiveness of the structural learning component RLC using (RLC + CN) connection-based model. These 2 methods are described below:

\begin{itemize}
    \item \textbf{ER + CN}: The ER component and the CN were trained and tested while being disconnected from the RLC component to prevent the structural learning part i.e., the size of the network does not change throughout the training process.
    \item \textbf{RLC + CN}: The RLC component and the CN were trained and tested with no data was being pushed into the buffer for replay.
\end{itemize}

In Task-incremental setting, it was observed that (ER + CN) connection-based model had a lower AF value compared to (RLC + CN) overall across the 2 datasets, while no such trend was observed in AA value. This has been detailed in Table \ref{tab:abl-task}.

\begin{table}[H]
\centering
\caption{Ablation study with Class-incremental setting on the components of GCL-GCN.}
\label{tab:abl-class}
\resizebox{\linewidth}{!}{%
\begin{tabular}{|c|cc|cc|} 
\hline
\multirow{2}{*}{\textbf{Class-incremental setting}} & \multicolumn{2}{c|}{Cora} & \multicolumn{2}{c|}{Citeseer} \\ 
\cline{2-5}
 & AA$(\uparrow)$ & AF$(\downarrow)$ & AA$(\uparrow)$ & AF$(\downarrow)$ \\ 
\hhline{|=====|}
ER + CN & $39.8\pm 9.1\%$ & $4.5\pm 27.7\%$ & $25.7\pm 9.7\%$ & $19.6\pm 26.9\%$ \\
RLC + CN & $32.4\pm 2.0\%$ & $42.8\pm 22.9\%$ & $29.7\pm 2.3\%$ & $38.7\pm 22.7\%$ \\ 
\hhline{|=====|}
GCL-GCN/ RLC + ER + CN & $55.7\pm 1.3\%$ & $2.5\pm 14.3\%$ & $36.5\pm 5.4\%$ & $2.8\pm 8.2\%$ \\
\hline
\end{tabular}
}
\end{table}

In Class-incremental learning, it is observed that overall (ER + CN) showed a better Backward Transfer or lower AF value in the 2 datasets. But in terms of Average Accuracy (AA), (RLC + CN) performed better on Citeseer dataset, while performing relatively poorly on Cora dataset as observed in Table \ref{tab:abl-class}.

% Need writing here

\section{Conclusion}
Our work presented a novel generalised framework for task-incremental and class-incremental setting with different variants of MP-GNNs. This framework combines the concept of structural learning that dynamically evolves the core Child Network (CN) hidden layers based on a trainable Reinforcement Learning based Controller (RLC) and an experience replay module for replaying the stored data and outputs from preceding tasks. For both task-incremental and class-incremental settings, the performance superiority of our method compared to the baselines increases as the size of the graph dataset increases.
Hence, we can conclude that our method performs better than the existing standards and is quite scalable across larger graphs. 

\begin{acks}
This work was supported in part by the Ministry of Education, Singapore, under its Academic Research Fund Tier 1, under Grant RG78/21.
\end{acks}

\bibliographystyle{ACM-Reference-Format}
\bibliography{sample-base}

\end{document}